\documentclass[10pt,twocolumn,letterpaper]{article}

\usepackage{cvpr}              % To produce the CAMERA-READY version
\usepackage[dvipsnames]{xcolor}

\definecolor{cvprblue}{rgb}{0.21,0.49,0.74}
\usepackage[pagebackref,breaklinks,colorlinks,citecolor=cvprblue]{hyperref}
\usepackage{multirow}
\usepackage{adjustbox}

\usepackage{xspace}
\makeatletter 
\DeclareRobustCommand\onedot{\futurelet\@let@token\@onedot} 
\def\@onedot{\ifx\@let@token.\else.\null\fi\xspace}  
\def\eg{\emph{e.g}\onedot} 
 
\def\ie{\emph{i.e}\onedot}

\makeatother 

\usepackage[T1]{fontenc}
\newcommand\blfootnote[1]{%
  \begingroup
  \renewcommand\thefootnote{}\footnote{#1}%
  \addtocounter{footnote}{-1}%
  \endgroup
}

\def\paperID{2932} % *** Enter the Paper ID here
\def\confName{CVPR}
\def\confYear{2026}

\title{STAGE: Storyboard-Anchored Generation for Cinematic Multi-shot Narrative}

\author{
    $\textrm{Peixuan Zhang}^{\#1} \quad \textrm{Zijian Jia}^{\#1} \quad \textrm{Kaiqi Liu}^{2,3} \quad \textrm{Shuchen Weng}^{*2,4} \quad \textrm{Si Li}^{*1} \quad  \textrm{Boxin Shi}^{2,3}$\\
    \small
    $\textrm{}^{1}$School of Artificial Intelligence, Beijing University of Posts and Telecommunications \\
    \small
    $\textrm{}^{2}$State Key Laboratory for Multimedia Information Processing, School of Computer Science, Peking University\\
    \small
    $\textrm{}^{3}$National Engineering Research Center of Visual Technology, School of Computer Science, Peking University \\
    \small
    $\textrm{}^{4}$Beijing Academy of Artificial Intelligence \\
    {\tt\footnotesize $\{$pxzhang,jiazijian,lisi$\}$@bupt.edu.cn}, {\tt\footnotesize  liukq04@gmail.com}, {\tt\footnotesize $\{$shuchenweng,shiboxin$\}$@pku.edu.cn}
}

\begin{document}
\twocolumn[{%
\renewcommand\twocolumn[1][]{#1}%
\maketitle

\begin{center}
    \centering
    \captionsetup{type=figure}
    \includegraphics[width=\linewidth]{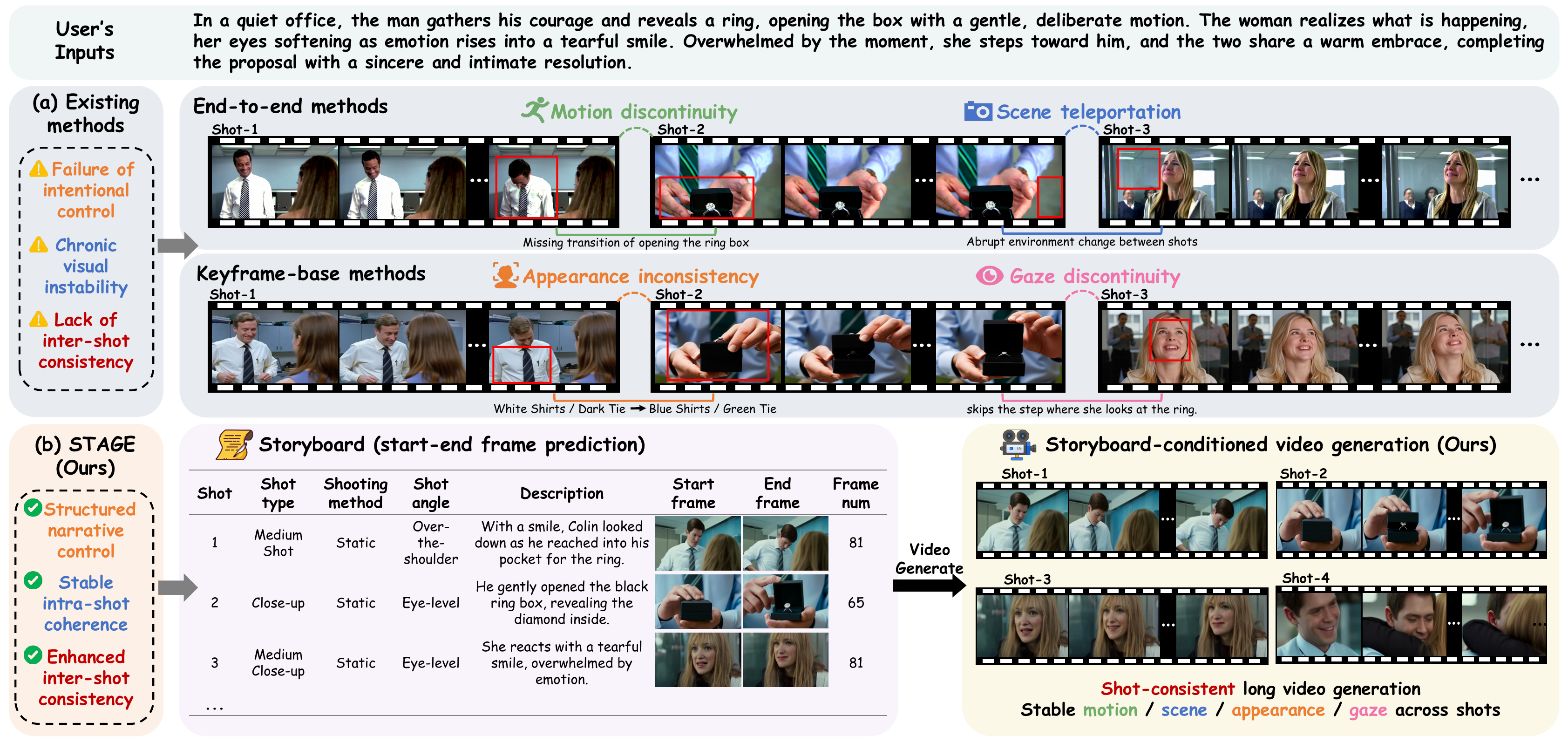}
  \captionof{figure}{Illustration of our proposed STAGE workflow for multi-shot video generation. Given a user-provided story description, existing end-to-end~\cite{cinetrans} and keyframe-based~\cite{iclora} methods often suffer from incoherent inter-shot transitions (\eg, motion discontinuities and appearance inconsistencies) that disrupt the narrative flow. We address this by explicitly predicting a structural storyboard composed of start-end frame pairs, which serve as visual anchors to ensure superior long-range shot consistency.}
    \label{fig:teaser}
\end{center}
}]

\begin{abstract}
\blfootnote{$^\#$ Equal contributions. * Corresponding author.}
While recent advancements in generative models have achieved remarkable visual fidelity in video synthesis, creating coherent multi-shot narratives remains a significant challenge. To address this, keyframe-based approaches have emerged as a promising alternative to computationally intensive end-to-end methods, offering the advantages of fine-grained control and greater efficiency. However, these methods often fail to maintain cross-shot consistency and capture cinematic language. In this paper, we introduce \textbf{STAGE}, a \textbf{ST}oryboard-\textbf{A}nchored \textbf{GE}neration workflow to reformulate the keyframe-based multi-shot video generation task. Instead of using sparse keyframes, we propose STEP$^2$ to predict a structural storyboard composed of start-end frame pairs for each shot. We introduce the multi-shot memory pack to ensure long-range entity consistency, the dual-encoding strategy for intra-shot coherence, and the two-stage training scheme to learn cinematic inter-shot transition. We also contribute the large-scale ConStoryBoard dataset, including high-quality movie clips with fine-grained annotations for story progression, cinematic attributes, and human preferences. Extensive experiments demonstrate that STAGE achieves superior performance in structured narrative control and cross-shot coherence. Our code will be available at \href{https://github.com/escapistmost/Storyboard-Anchored-Generation}{this url}.

\end{abstract}

\section{Introduction}

Recent advancements in generative models have dramatically improved the visual quality of synthesized videos, fueling their adoption on short-clip content platforms~\cite{sora,sora2, kling}. To further unlock their potential for more sophisticated applications (\eg, filmmaking), recent models are evolving to enhance storytelling ability~\cite{storydiffusion, seedstory,synthesizing}. However, this narrative ability typically requires composing a long video from multiple distinct shots depicting different stages or perspectives of a story, which still poses significant challenges for existing models.

While directly generating an entire multi-shot video in an end-to-end manner~\cite{cinetrans,storyanchors,shotadapter,streamingt2v} is an intuitive approach, it is computationally expensive and runs in an ``all-or-nothing'' paradigm, leading to an inefficient trial-and-error process with limited user control. To provide fine-grained user control, keyframe-based approaches~\cite{cut2next,movieagent,videogenofthought,moviedreamer} first generate several keyframes to establish the video's narrative structure, and subsequently employ external I2V models~\cite{dynamicrafter,gen4,svd} to synthesize each shot accordingly. However, these methods suffer from disrupted cross-shot coherence and struggle to capture cinematic language (\cref{fig:teaser} (a), the close-up reveals motion discontinuity when opening the ring box).

In this paper, we reformulate the keyframe-based approach, proposing a structural storyboard composed of Start-End frame pairs defining each shot (\ie, $(F_1^\mathrm{S}, F_1^\mathrm{E}), (F_2^\mathrm{S}, F_2^\mathrm{E}), ...$), instead of a sparse keyframe sequence.
This formulation offers three advantages:
\textit{(i) Structured narrative control:} These start-end frame pairs form a robust narrative scaffold, ensuring long-range consistency of entities (\eg, character appearance) and settings (\eg, scene background) throughout the entire video.
\textit{(ii) Intra-shot coherence:} The intra-shot pair $(F_i^\mathrm{S}, F_i^\mathrm{E})$
explicitly anchors visual content (\eg, character) and the intended progression within the shot (\eg, camera movements).
\textit{(iii) Inter-shot transitions:} The inter-shot pair $(F_i^\mathrm{E}, F_{i+1}^\mathrm{S})$ explicitly models the shot transition, effectively conveying complex cinematic language (\eg, shot/reverse shot).
However, realizing these intra-, inter-, and global advantages is non-trivial, demanding both a tailored model and a structurally annotated dataset.

To address this challenge, we introduce \textbf{STAGE}, a workflow for \textbf{ST}oryboard-\textbf{A}nchored \textbf{GE}neration for cinematic multi-shot narrative.
The technical core of this workflow is our \textbf{ST}art-\textbf{E}nd frame-\textbf{P}air \textbf{P}rediction model (\textbf{STEP$^2$}), which iteratively visualizes the text-based storyboard into $(F_i^\mathrm{S}, F_i^\mathrm{E})$ pairs.
To further enable STEP$^2$ to visualize the storyboard of each shot effectively, we propose: 
\textit{(i)} a multi-shot memory pack to compress the history and ensure long-range entity consistency;
\textit{(ii)} a dual-encoding strategy to enforce intra-shot coherence and logical correlation; and
\textit{(iii)} a two-stage training scheme to understand the complex cinematic language of inter-shot transitions.
In the complete STAGE workflow, a director agent first augments the user-provided story theme into a text-based storyboard. Our fully-trained STEP$^2$ model then iteratively generates the $(F_i^\mathrm{S}, F_i^\mathrm{E})$ pairs. These pairs are finally fed into an off-the-shelf video generation model (\eg, WanX~\cite{wanx}, Veo3.1~\cite{veo3}) to generate video clips, which are then concatenated into the final multi-shot video.

To facilitate the training of STEP$^2$, we collect open-sourced movie clips~\cite{condensedmovie} and construct the ConStoryBoard dataset\footnote{We will publish the dataset once the paper is accepted.}. It consists of 100K video clips with start-end frame pairs, along with structural storyboard annotations (\eg, story progress of each shot) and diverse cinematic language attributes (\eg, shot scale, shot length, camera angle, and camera movement). Building on this, we further curate the ConStoryBoard-HP by manually selecting high-quality frame pairs and constructing preference tuples for human-preference alignment post-training.

We conclude our contributions as follows:
\begin{itemize}
    \item We reformulate the keyframe-based approach by introducing STEP$^2$, which predicts structural start-end frame pairs to enable controllable multi-shot video generation.
        
    \item We design a dual-encoding strategy to ensure logical correlation within each shot, and a multi-shot memory pack to provide long-range consistency of entities.
    
    \item We construct the ConStoryBoard dataset with fine-grained annotations, including a manually selected preference subset to optimize inter-shot cinematic language.

\end{itemize}

\section{Related work}

\subsection{Single-Shot video generation}
The field of single-shot video generation has seen remarkable progress. Building on the success of image diffusion models~\cite{stablediffusion,imagen,sdxl,pixart}, early methods~\cite{imagenvideo,makeavideo,vdm} adapt pre-trained models by incorporating temporal layers. However, these approaches often result in limited temporal coherence. To address these limitations, research has shifted towards Diffusion Transformer architectures~\cite{scalable}. While state-of-the-art video foundation models~\cite{cogvideox,hunyuanvideo,wanx} are able to generate high-fidelity videos with a single shot, they typically lack a native mechanism for multi-shot narratives. Consequently,  naively applying them shot-by-shot results in jarring visual discontinuities that disrupt the narrative flow, marking the key challenge our work addresses.

\subsection{Multi-Shot video generation}
Generating multi-shot videos requires maintaining both content consistency and narrative coherence across shots. Existing methods extend single-shot models to generate full sequences, either via end-to-end visual conditioning~\cite{videostudio, moviedreamer} or by modifying attention mechanisms~\cite{shotadapter, cinetrans, holocine}. These approaches are computationally expensive and offer limited user control. To address these limitations, an alternative approach~\cite{storydiffusion, seedstory, movieagent, cineverse} first generates sparse keyframes and then synthesizes the full video clips based on them. Despite their advances in text fidelity and story coherence, they largely neglect the cinematic language for inter-shot transitions, leading to abrupt cuts and logical disconnects that break the narrative flow.
In contrast, our STAGE explicitly reformulates the keyframe-based approach as a start-end frame pair prediction problem, thereby directly modeling these inter-shot transitions.

\subsection{Reinforcement learning for visual generation}
Inspired by the success of Reinforcement Learning from Human Feedback (RLHF) in language models, aligning visual models with human preferences is a promising research area. 
Initial efforts involve directly fine-tuning models with scalar reward signals~\cite{imagereward, clark2023directly} or employing reward weighted regression~\cite{peng2019advantage,lee2023aligning}. Subsequently, policy gradient algorithms (\eg, PPO~\cite{ppo}) are integrated into diffusion models, demonstrating notable efficacy in visual quality improvement~\cite{fan2023reinforcement,miao2024training}.
To avoid significant computational overhead and training instability, Direct Preference Optimization (DPO)~\cite{dpo} emerges and demonstrates its effectiveness in aligning visual models with human preferences for static image properties, such as text-image fidelity~\cite{wallace2024diffusion,wang2025diffusion} and aesthetic quality~\cite{liang2025aesthetic}.
Building on this, we extend this technique to optimize for the complex temporal relationships (\eg, cinematic language) that define coherent storytelling in multi-shot videos.

\section{Dataset} \label{sec:dataset}
Current keyframe-based video datasets~\cite{cinetrans, storyboard, cineverse} offer a single keyframe for each shot, primarily focusing on text-to-shot alignment while  neglecting the start and end frames for modeling inter-shot transitions.
This motivates us to construct the ConStoryBoard dataset, tailored to train our start-end frame pair prediction model.

The construction of ConStoryBoard begins with multi-shot videos collected from the Condensed Movies dataset~\cite{condensedmovie}, which we filter for high-resolution (over 1080p) and high aesthetic scores~\cite{laion} (over 5.5). These filtered videos are then segmented into individual video clips for each shot using TransNetV2~\cite{transnetv2}. For each resulting clip ($i$-th shot in the original video), we employ InternVL-3.5~\cite{internvl} to generate a detailed text description $D_i$ of the story progression, and structured cinematic attributes $C_i$ (\eg, shot scale and camera movement).
We then extract the ground-truth start-end frame pair $(F^\mathrm{S}_i, F^\mathrm{E}_i)$ from these annotated clips, and post-process them by cropping black bars and removing watermarks.
Finally, the ConStoryBoard dataset comprises 100K training pairs and 1K testing pairs, where each sample consists of a ground-truth pair $(F^\mathrm{S}_i, F^\mathrm{E}_i)$ and its corresponding annotations $(D_i, C_i)$.

To facilitate model alignment with human preferences, we further curate ConStoryBoard-HP (Human-Preferred) by manually selecting the most high-quality and cinematically significant pairs from our dataset.
To construct the preference tuples, we define the ground-truth pair $(F_i^\mathrm{S}, F_i^\mathrm{E})$ as the positive sample $y_w$. 
Observing that internal frames from the same video clip typically represent a mismatched cinematic language (\eg, an incomplete camera movement), we randomly sample two internal frames from the same clip as the corresponding negative sample $y_l$.
These preference tuples $(y_w, y_l)$ are used for the alignment post-training.

\begin{figure}[t]
  \centering
  \includegraphics[width=\linewidth]{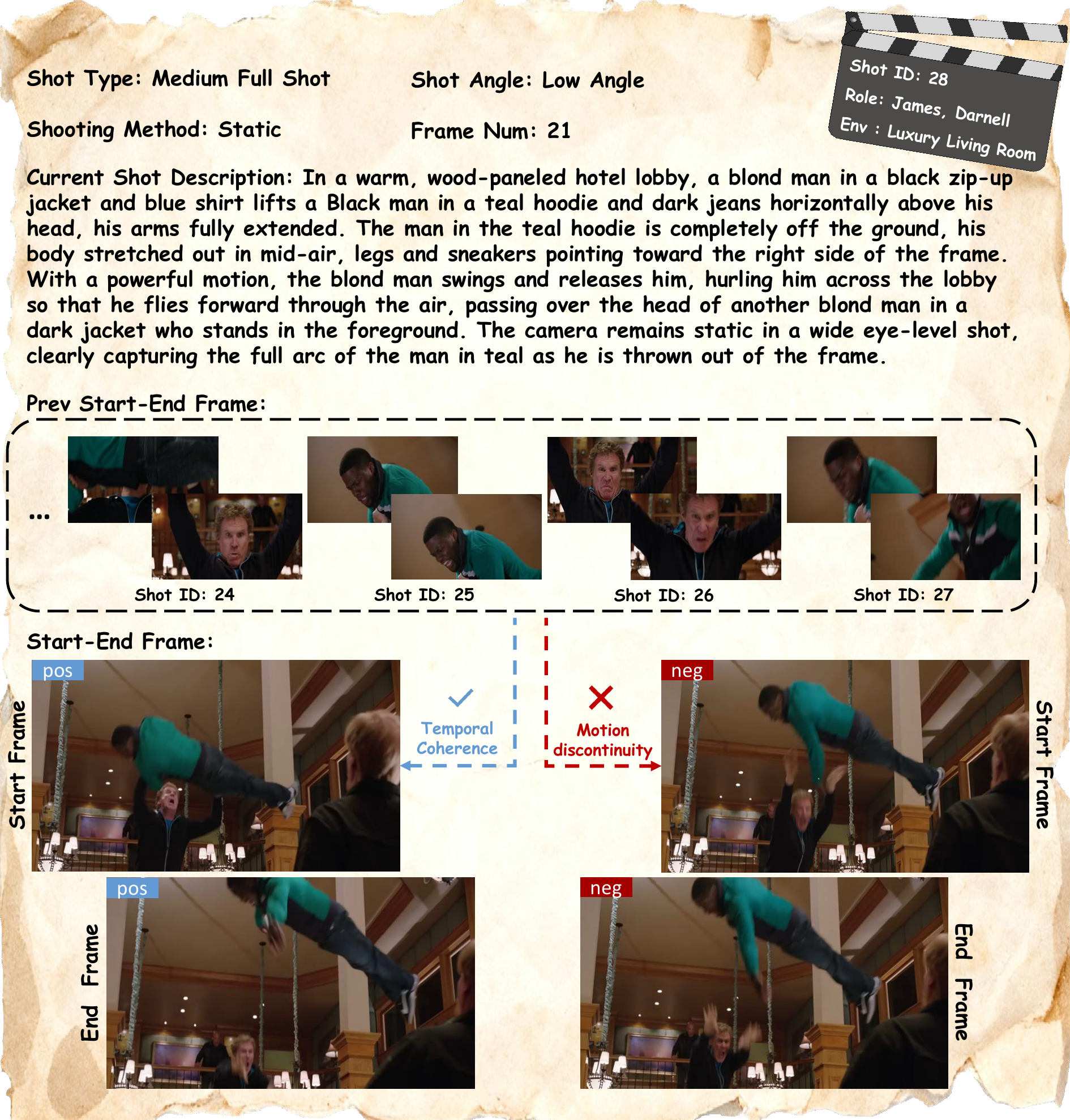}
  \caption{An example of the ConStoryBoard dataset.
  }
  \label{fig:dataset}
\vspace{-2mm}
\end{figure}

\begin{figure*}[t]
   \centering
  \includegraphics[width=\linewidth]{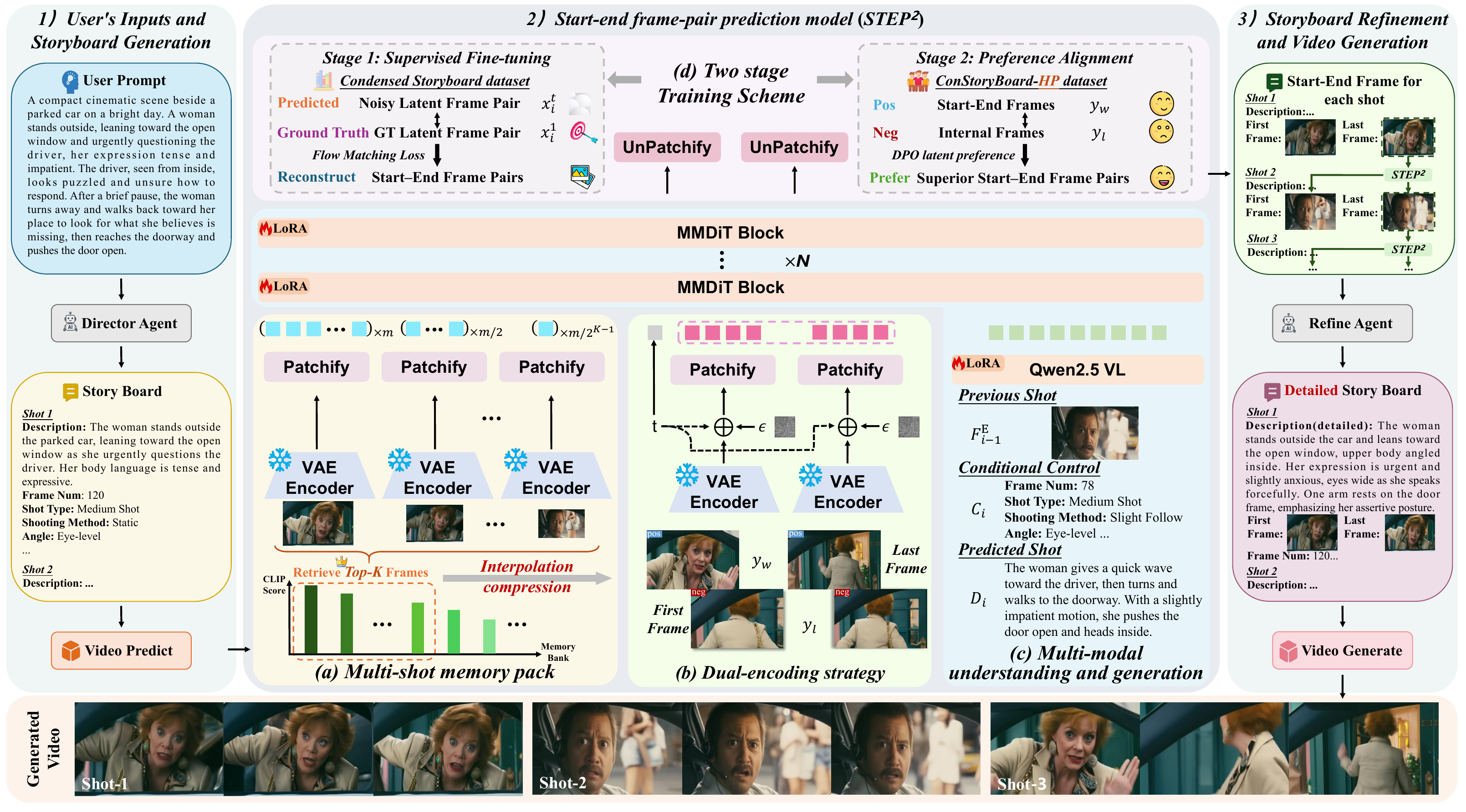}
  \caption{Overview of our proposed STAGE workflow.   
  The core component of STAGE is the \textbf{Start-End frame-pair prediction model (STEP$^2$)}, which iteratively generates start-end frame pairs for each shot.
  To ensure long-term consistency, STEP$^2$ is equipped with the \textbf{(a) multi-shot memory pack} that compresses preceding visual context into a compact token. The \textbf{(b) dual-encoding strategy} is adopted to maintain intra-shot coherence by jointly encoding the start and end frames of the current shot.
  By integrating \textbf{(c) multi-modal understanding and generation} models into a unified architecture, STEP$^2$ performs robust reasoning from diverse contexts to ensure inter-shot coherence (\cref{sec:step2}). 
  STEP$^2$ is optimized via a \textbf{(d) two-stage training scheme}: an initial supervised fine-tuning stage establishes a strong generative foundation, followed by a preference alignment stage to align outputs with human preferences (\cref{sec:twostage}).
  During inference, STEP$^2$ is integrated into the STAGE workflow. A Director Agent first generates a structured storyboard from a user's theme. Then, STEP$^2$ produces the corresponding start-end frame pairs for each shot. Finally, a Refiner Agent uses these frames to create enhanced prompts that guide an off-the-shelf video model to synthesize the final multi-shot video (\cref{sec:stage_workflow}).
  }
  \label{fig:pipeline}
  \vspace{-2mm}
\end{figure*}

\section{Method}
This section begins with an introduction to the Start-End frame-pair prediction model (STEP$^2$) to iteratively generate start-end frame pairs (Sec.~\ref{sec:step2}). 
Then, we design the two-stage training scheme to train the STEP$^2$ model for understanding cinematic language (Sec.~\ref{sec:twostage}). 
Finally, we present the STAGE workflow to demonstrate the approach of STEP$^2$ applied for multi-shot video generation (Sec.~\ref{sec:stage_workflow}).

\subsection{Start-end frame pair prediction model} \label{sec:step2}

\noindent \textbf{Multi-shot memory pack.}
To enable iterative generation of start-end frame pairs with long-term temporal consistency of entities, the STEP$^2$ requires a memory mechanism for preceding shots. 
Since referring to all previous shots will bring a remarkable computational burden, we propose the multi-shot memory pack to compress these shots into a compact context. 

Specifically, as illustrated in \cref{fig:pipeline} (a), given preceding shot start-end frame pairs $\{(F_j^\mathrm{S}, F_j^\mathrm{E})\}_{j=1}^{i-1}$ for $i$-th shot, we disorderly collect all frames in these pairs into a memory bank $\{F_j^\mathrm{M}\}_{j=1}^{2i-2}$, and encode all these frames into the latent space $\{m_j\}_{j=1}^{2i-2} = \{\mathcal{E}_\mathrm{vae}(F_j^\mathrm{M})\}_{j=1}^{2i-2}$ using a pre-trained VAE encoder $\mathcal{E}_\mathrm{vae}$.
To guide the iterative generation with semantically relevant memory cues, we rank these preceding latent codes based on their CLIP similarity~\cite{clip} as $\{m'_{j}\}_{j=1}^{2i-2} = \mathrm{Sort}\big(\{m_j\}_{j=1}^{2i-2}\big)$.
After that, this ranked memory bank is further compressed into a packed memory token via a progressive spatial tiling mechanism:
\begin{equation} \label{eq:memory_pack_area} 
    M_i = \mathrm{SpatialTile}_{j \in \{1,\dots,2i-2\}} \big(\mathcal{P}(m'_{j}, A_j)\big),
\end{equation}
where $\mathcal{P}(m'_{j}, A_j)$ is the downsample function that compresses the latent code $m'_j$ based on the compression rate $A_j = \frac{1}{2^{j}}$ to ensure the total area for $M_i$ is mathematically converged  (\ie, $\sum_{j=1}^{\infty} A_j$=1).
These packed memory tokens $M_i$ are then fed into the generation model $\mathcal{E}_{gen}$ to present a potentially infinite generation history.

\noindent \textbf{Dual-encoding strategy.}
To ensure the predicted start-end frame pair maintains intra-shot story coherence (\eg, scenes remain visually consistent) and temporal dynamics (\eg, a zoom-in cinematic language), we propose a dual-encoding strategy to enable the implicit visual context sharing between the start and end frames.

Specifically, as shown in \cref{fig:pipeline} (b), we encode the ground truth start-end frame pair $(F_i^\mathrm{S}, F_i^\mathrm{E})$ separately using a pre-trained VAE encoder $\mathcal{E}_{vae}$, and then concatenate them along the sequence dimension to construct a single joint shot tensor $x_{i} = \big[\mathcal{E}_\mathrm{vae}(F_i^\mathrm{S}); \mathcal{E}_\mathrm{vae}(F_i^\mathrm{E})\big]$.
As the common practice of flow matching~\cite{lipman2022flow}, this shot tensor is then linearly interpolated with a Gaussian noise as:
\begin{equation}\label{eq:flow_matching_noise}
    x_{i}^t = t \cdot x_{i}^1 +  (1 - t) \cdot x_{i}^0,
\end{equation}
where $t \in [0, 1]$ is the continuous time step, $x_i^1 = x_{i}$ is the clean shot tensor, and $x_i^0 \sim \mathcal{N}(0, \mathbf{I})$ is Gaussian noise.

\noindent \textbf{Multi-modal understanding and generation.}
To reason about shot semantics from diverse context, our STEP$^2$ first adopts a multi-modal large language model to understand the story progress and character performance of each shot. Specifically, as illustrated in \cref{fig:pipeline} (c), the understanding model $\mathcal{E}_{mu}$ is built upon Qwen2.5-VL~\cite{qwen2.5vl}, which leverages the end frame of the previous shot $F_{i-1}^\mathrm{E}$, the $i$-th shot corresponding text description $D_i$, and cinematic attributes $C_i$ to generate unified context tokens $U_i = \mathcal{E}_{mu}(F_{i-1}^\mathrm{E}, D_i, C_i)$ for the $i$-th shot. 

With the aforementioned memory tokens $M_i$ and interpolated joint shot tensor $x_i^t$, these unified context tokens $U_i$ are fed into the generation model $\mathcal{E}_{gen}$ to generate the corresponding start-end frame pair $(F_i^\mathrm{S}, F_i^\mathrm{E})$. Specifically,  the generation model $\mathcal{E}_{gen}$ is implemented as a diffusion Transformer architecture, comprising multiple MMDiT blocks~\cite{esser2024scaling} that includes the self-attention layers for global context interaction. This enables our STEP$^2$ to generate clean start-end frame pairs referring all integrated and compression semantics by solving the ODE:
\begin{equation} \label{eq-ode}
    \textrm{d}x_i^t/\textrm{d}t = \mathcal{E}_\mathrm{gen}(U_i, t, x_i^t,  M_i),
\end{equation}
from $t = 0$ to $t = 1$ using a numerical solver.

\subsection{Two-stage training scheme} \label{sec:twostage}

\noindent \textbf{Supervised fine-tuning.}
We first conduct a supervised fine-tuning (SFT) on our ConStoryBoard dataset to pre-train the STEP$^2$ model, establishing a strong foundation for inter-shot transitions.
Specifically, we employ the Low-Rank Adaptation (LoRA) technique~\cite{lora} on both the understanding model $\mathcal{E}_{mu}$ (for high-level semantic understanding) and the generation model $\mathcal{E}_{gen}$ (for low-level frame generation).
Following the flow matching formulation~\cite{liu2022flow}, our training target is the constant velocity vector field $v_t = x_i^1 - x_i^0$, pointing from the noise $x_i^0 \sim \mathcal{N}(0, \mathbf{I})$ to the ground-truth latent pair $x_i^1$:
\begin{equation}
\label{eq:flow_matching}
\mathcal{L}_{\text{SFT}}  =  \mathbb{E}_{x_i^{1}, x_i^0, \mathcal{C}_i, t} \| v_\theta(x_i^{t}, t, \mathcal{C}_i) - v_t \|^2,
\end{equation}
where $\mathcal{C}_i = [D_i, C_i, \{(F_j^\mathrm{S}, F_j^\mathrm{E})\}_{j=1}^{i-1}]$ represents  the text description, the cinematic attribute, and all the preceding start-end frame pairs and $v_\theta$ is our entire STEP$^2$ model.

\noindent \textbf{Preference alignment.}
To further align our STEP$^2$ model with human preferences, we perform the post-training using Direct Preference Optimization (DPO)~\cite{dpo}. Using the SFT-trained model as the reference model $v_\text{ref}$ and extracting preference tuple $(y_w, y_l)$ from the {ConStoryBoard-HP} subset, this process aims to maximize the likelihood of policy model $v_\theta$  prefers $y_w$ over $y_l$:
\begin{equation}
\label{eq:dpo_loss}
\mathcal{L}_{\text{DPO}} = -\mathbb{E}_{(y_w, y_l), \mathcal{C}_i, t} \left[ \log \sigma \left( \beta (\text{D}_\theta - \text{D}_\text{ref}) \right) \right],
\end{equation}
where $\sigma(\cdot)$ is the sigmoid function and $\beta$ is a scaling parameter. The preference differences for the policy and reference models are defined as:
% \begin{align}
\begin{equation}
\text{D}_k\! =\! \| v_k(\hat{x}_i^{t}, t, \mathcal{C}_i) - \hat{v}^{t} \|^2 - \| v_k(\check{x}_i^{t}, t, \mathcal{C}_i) - \check{v}^{t} \|^2, 
\end{equation}
where $k \in \{\theta, \text{ref}\}$, $\hat{x}_i^{t}$ and $\check{x}_i^{t}$ are noisy latent codes from the negative $y_l$ and positive sample $y_w$, respectively.
% and $\hat{v}_t$ and $\check{v}_t$ is the ground-truth velocity $x_{w}^1 - x_i^0$ .
The $\hat{v}^{t}$ and $\check{v}^{t}$ are their corresponding ground-truth velocity vectors, defined as $\hat{v}^{t} = x_{i,l}^1 - x_{i,l}^0$ and $\check{v}^{t} = x_{i,w}^1 - x_{i,w}^0$. Here, $x_{i,l}^1$ and $x_{i,w}^1$ are the VAE-encoded latents for $y_l$ and $y_w$.

\subsection{STAGE workflow} \label{sec:stage_workflow}
We propose the STAGE workflow to generate coherent multi-shot videos $V=[V_1, \dots, V_N]$ based on a story description $T_\mathrm{desc}$, where our well-trained  STEP$^2$ model is integrated to predict start-end frame pairs for each shot. This workflow is organized into three main stages:

\begin{figure*}[t]
  \centering
  \includegraphics[width=\linewidth]{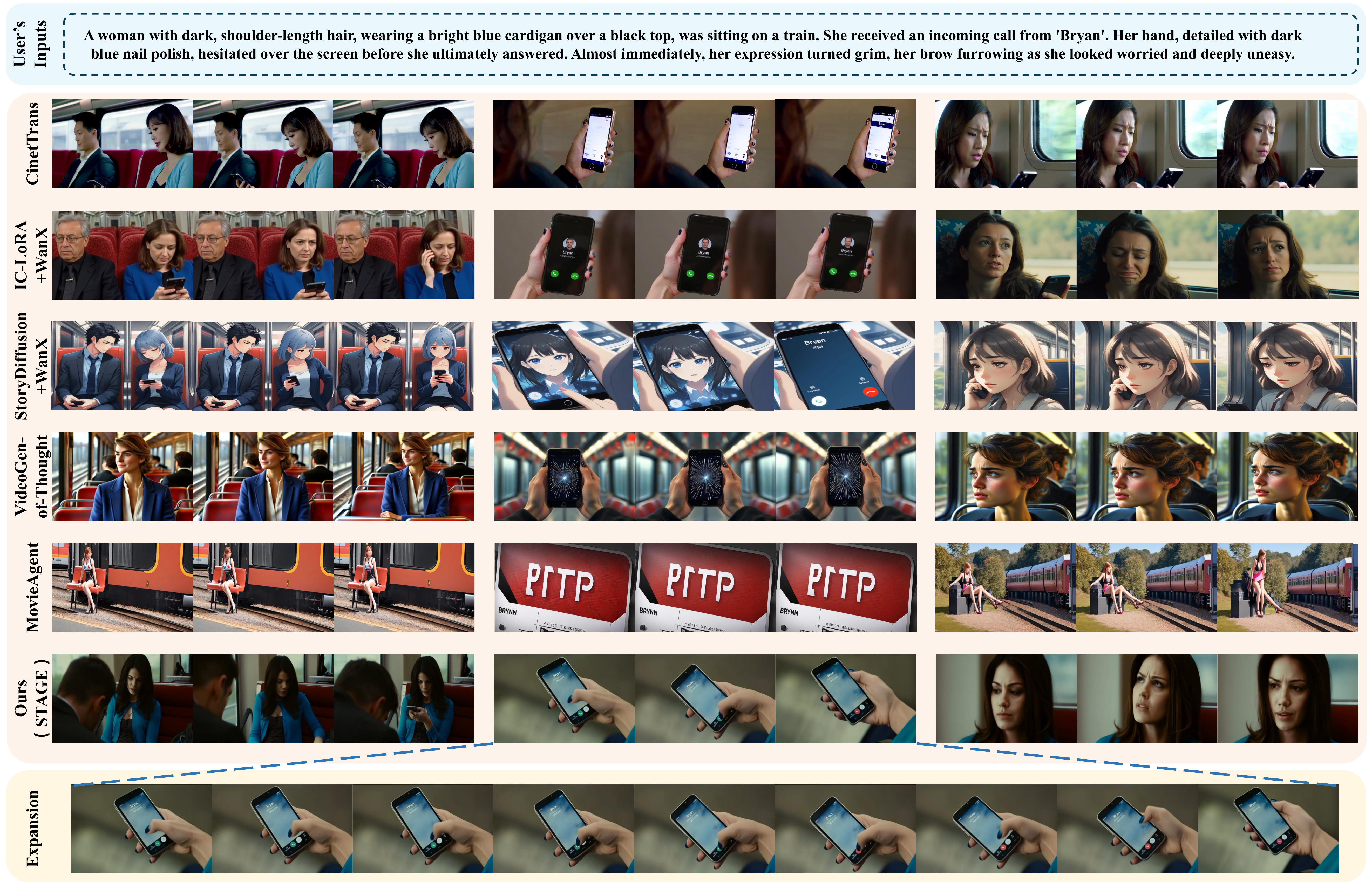}
  \caption{Visual quality comparisons with multi-shot video generation methods.}
  \label{fig:comparison}
  \vspace{-2mm}
\end{figure*}

\textit{1) User's Inputs and Storyboard Generation:}
Given a user-provided text-based story theme $T_{\text{desc}}$, we design a Director Agent $G_{\text{dire}}$ with chain-of-thought~\cite{cot} pre-defined prompts to generate a structured storyboard $\mathcal{S}$:
\begin{equation}
    \label{eq:storyboard_generation}
    \mathcal{S} = \mathrm{G}_{\text{dire}}(T_{\text{desc}}),
\end{equation}
where $\mathcal{S} = \{D_i, C_i\}_{i=1}^N$ is a sequence of $N$ shot specifications $\mathcal{S}_i = (D_i, C_i)$, each one details the shot of story progress with a text description $D_i$ and a set of cinematic attributes $C_i$.

\textit{2) Iterative Start-End Frame Generation:}
Leveraging the well-trained Start-End frame-Pair Prediction model (\ie, STEP$^2$),  the abstract storyboard for each shot is iteratively translated into concrete visual anchors (\ie, a start-end frame pair $(F_{i}^\mathrm{S}, F_{i}^\mathrm{E})$):
\begin{equation}
\label{eq:step2_generation}
    (F_i^\mathrm{S}, F_i^\mathrm{E}) = \text{STEP}^2(D_i, C_i, \{(F_j^\mathrm{S}, F_j^\mathrm{E})\}_{j=1}^{i-1}).
\end{equation}
Since the first shot ($i=1$) has no preceding context, the model is conditioned solely on its specification $(D_i, C_i)$. This process iteratively generates a complete sequence of frame pairs to the last shot ($i=N$).

\textit{3) Storyboard Refinement and Video Generation:}
To bridge the gap between static keyframes and dynamic video, we employ a Shot Refiner Agent $G_{\text{refiner}}$. This agent integrates the original storyboard specification $(D_i, C_i)$ with the generated visual anchors $(F_i^\mathrm{S}, F_i^\mathrm{E})$ to produce a comprehensive text description $R_{i}$ for each shot:
\begin{equation}
    \label{eq:shot_refiner}
    R_{i} = G_{\text{refiner}}(D_i, C_i, F_i^\mathrm{S}, F_i^\mathrm{E}).
\end{equation}
Subsequently, this refined description $R_{i}$ and the start-end frame pair $(F^\mathrm{S}_i, F^\mathrm{E}_i)$ are fed into an off-the-shelf video generation model $G_{\text{video}}$ (\eg, WanX~\cite{wanx}, Veo3.1~\cite{veo3}) to guide the generation process of the video clip $V_i$ of the $i$-th shot:
\begin{equation}
    \label{eq:video_synthesis}
    V_i = G_{\text{video}}(R_{i}, F_i^\mathrm{S}, F_i^\mathrm{E}).
\end{equation}
Finally, the multi-shot narrative video $V=[V_1, \dots, V_N]$ is 
produced by concatenating all individual clips.

\vspace{-2mm}
\section{Experiment}
% \vspace{-1mm}
\subsection{Training details} \label{sec:detail}
Our implementation is based on Qwen-Image~\cite{qwenimage} where VAE is frozen. The rank of LoRA weight is set to 64.
Training is performed on 8 A800 GPUs using the Adam optimizer \cite{kingma2014adam} at a learning rate of $1 \times 10^{-4}$.
Our two-stage training scheme consists of an initial 100K iterations of supervised fine-tuning, followed by an additional 20K iterations of preference alignment.

\subsection{Quantitative evaluation metrics}
We comprehensively evaluate STAGE across five aspects\footnote{Metric details are provided in the supplementary materials.}:
\textit{(i)} Overall video quality. We assess the perceptual quality of generated videos using the \textbf{Aesthetic Quality (AQ)} and \textbf{Image Quality (IQ)} metrics from VBench~\cite{vbench}.
\textit{(ii)} Text video consistency. To measure the alignment between the generated video and the input text description, we employ the \textbf{Overall Consistency (OC)} metric from VBench~\cite{vbench}.
\textit{(iii)} Intra-shot coherence. We evaluate the temporal consistency within individual shots using \textbf{Subject Consistency (SC)} and \textbf{Background Consistency (BC)} from VBench~\cite{vbench}.
\textit{(iv)} Inter-shot coherence. We extend VBench's intra-shot metrics for inter-shot evaluation, defining \textbf{Subject Consistency-Extra (SC-E)} and \textbf{Background Consistency-Extra (BC-E)} metrics.
\textit{(v)} Inter-shot transition. We introduce \textbf{Transition Vector Similarity (TVS)} to measure cinematic transition quality.
To further evaluate the quality of multi-shot video, we employ the VLM~\cite{gemini} to rate \textbf{Overall Video Quality (OVQ)}, \textbf{Video Text Consistency (VTC)}, \textbf{Inter-Shot Consistency (ISC)}, \textbf{Shot Transition Smoothness (STS)} on a scale of 0 to 1 score.

\begin{table*}[t]
% \vspace{-2mm}
\caption{Quantitative experiment results of comparison and ablation. $\uparrow$ ($\downarrow$) means higher (lower) is better. Throughout the paper, the best performances are highlighted in \textbf{bold}.} 
\vspace{-4mm}
\begin{center}
{
    \setlength\tabcolsep{6pt}
    \centering
    \begin{adjustbox}{width={\textwidth},totalheight={\textheight},keepaspectratio}
    \begin{tabular}{l | c c c c c c c c | c c c c}  \toprule
    \multirow{2}{*}{Method} & \multicolumn{8}{c|}{Quantitative metrics evaluation} & \multicolumn{4}{c}{LLM evaluation} \\
    & AQ $\uparrow$ & IQ $\uparrow$  & OC $\uparrow$  & SC $\uparrow$ & BC $\uparrow$ & SC-E $\uparrow$ & BC-E $\uparrow$  & TVS $\uparrow$ & OVQ $\uparrow$  & VTC $\uparrow$  & ISC $\uparrow$ & STS $\uparrow$  \\ \midrule
    \multicolumn{12}{c}{Comparison with state-of-the-art methods} \cr \midrule
        CineTrans~\cite{cinetrans} & \text{0.5652} & \text{0.6120} & \text{0.2018} & \text{0.9437} & \text{0.9504} & \text{0.6197} & \text{0.7428} & \text{0.0455} & \text{0.7972} & \text{0.3551} & \text{0.5585} & \text{0.4931}
        \\
        
        IC-LoRA~\cite{iclora} + WanX~\cite{wanx} & \text{0.6333} & \text{0.6951} & \text{0.2140} & \text{0.9567} &\text{0.9615} & \text{0.5319} & \text{0.7438} & \text{0.2090} & \text{0.7597} & \text{0.3897} & \text{0.4901} & \text{0.4696}
        \\

        StoryDiffusion~\cite{storydiffusion} + WanX~\cite{wanx} & \text{0.6941} & \text{0.7018} & \text{0.2087} & \text{0.9456} &\text{0.9640} & \text{0.5780} & \text{0.7988} & \text{0.1441} & \text{0.5343} & \text{0.2069} & \text{0.4813} & \text{0.4575}
        \\
        
        VideoGen-of-Thought~\cite{videogenofthought}  & \text{0.7210} & \text{0.6630} & \text{0.1689} & \text{0.9599} & \text{0.9554} & \text{0.6278} & \text{0.7830} & \text{0.0966} & \text{0.8106} & \text{0.1120} & \text{0.5086} & \text{0.4507}
        \\ 
         
        MovieAgent~\cite{movieagent} & \text{0.5742} & \text{0.7069} & \text{0.0711} & \text{0.9664} & \text{0.9384} & \text{0.4993} & \text{0.6473} & \text{0.0079} & \text{0.4895} & \text{0.1931} & \text{0.4511} & \text{0.4182} 
        \\ 
        Ours (STAGE) & \textbf{0.7689} & \textbf{0.7305} & \textbf{0.2713} & \textbf{0.9695} & \textbf{0.9685} & \textbf{0.6917} & \textbf{0.8207} & \textbf{0.2732} & \textbf{0.8929} & \textbf{0.6069} & \textbf{0.6985} & \textbf{0.6255}   \\ \midrule
    \multicolumn{12}{c}{Ablation study} \cr \midrule
        \textit{W/o} MMP & \text{0.7344} & \text{0.7220} & \text{0.2143} & \text{0.9631} & \text{0.9592} & \text{0.6088} & \text{0.7311} & \text{0.2370} & \text{0.8835} & \text{0.3642} & \text{0.5466} & \text{0.5228} \cr
        \textit{W/o} DES  & \text{0.7217} & \text{0.7145} & \text{0.2488} & \text{0.9542} & \text{0.9476} & \text{0.6803} & \text{0.8124} & \text{0.2680} & \text{0.7803} & \text{0.5063} & \text{0.6552} & \text{0.5167} \cr
        \textit{W/o} TTS &  \text{0.7476} & \text{0.7240} & \text{0.2635} & \text{0.9613} & \text{0.9633} & \text{0.6636} & \text{0.8037} & \text{0.2195} & \text{0.8733} & \text{0.5766} & \text{0.6614} & \text{0.5111} \cr \bottomrule
    
    \end{tabular}\label{tab:comparison}
    \end{adjustbox}
}
\end{center}
\vspace{-4mm}
\end{table*}

\begin{figure*}[t]
  \centering
  \includegraphics[width=\linewidth]{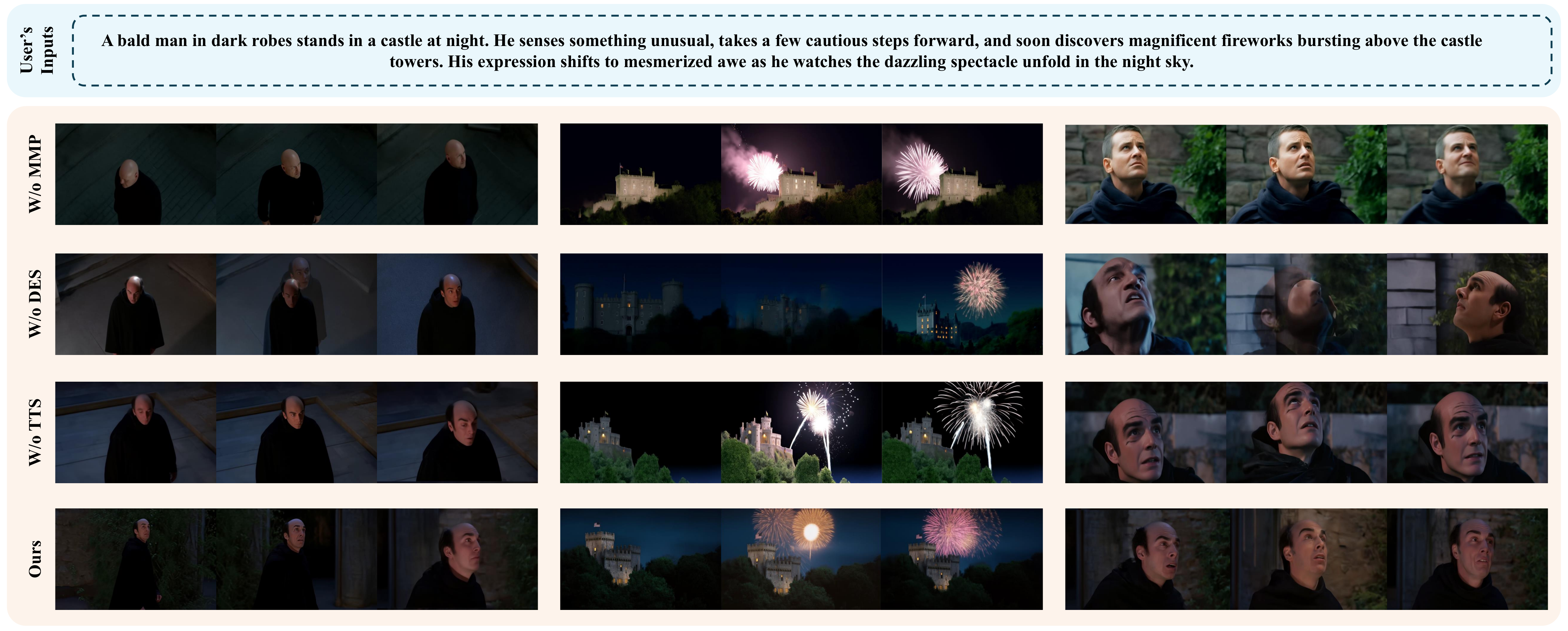}
  \caption{Ablation study results with different variants of our STAGE framework.}
  \label{fig:ablation}
  \vspace{-4mm}
\end{figure*}

\subsection{Comparison with State-of-the-art Methods} \label{sec:comparison}
We compare our approach with state-of-the-art multi-shot video generation methods, including end-to-end approaches (\ie, CineTrans~\cite{cinetrans}) and key-frame-based approaches (\ie, IC-LoRA~\cite{iclora} + WanX~\cite{wanx}, StoryDiffusion~\cite{storydiffusion} + WanX~\cite{wanx}, MovieAgent~\cite{movieagent}, and VideoGen-of-Thought~\cite{videogenofthought}).

\noindent \textbf{Qualitative comparisons.}
In Fig.~\ref{fig:comparison}, we present visual quality comparison results with aforementioned methods. 
Both CineTrans~\cite{cinetrans} and IC-LoRA~\cite{iclora} + WanX~\cite{wanx} fail to maintain consistency across multiple shots (\eg, \cref{fig:comparison} first and second row, the appearance of the train car window and the surrounding environment in the first shot are inconsistent with those in the third, leading to a disjointed narrative). 
StoryDiffusion~\cite{storydiffusion} + WanX~\cite{wanx} produces an overly cartoonish style and lower overall quality (\eg, \cref{fig:comparison} third row, the entire sequence exhibits a stylized aesthetic that is unsuitable for cinematic applications). 
VideoGen-of-Thought~\cite{videogenofthought} suffers from poor temporal coherence in its action sequences (\eg, \cref{fig:comparison} fourth row, the woman's hands are on her arm in the first shot but abruptly holds a phone with hands in the second, breaking the continuity of motion). 
MovieAgent~\cite{movieagent} exhibits low text fidelity (\eg, \cref{fig:comparison} fifth row, despite the prompt is sitting on a train, the generated video depicts a woman beside the train, failing to accurately reflect the user's intent). 
In contrast, our STAGE workflow successfully maintains high inter-shot consistency while ensuring shot smooth transitions, resulting in a coherent and visually superior cinematic result.

\noindent \textbf{Quantitative comparisons.}
We present quantitative comparisons in \cref{tab:comparison}, where our method significantly outperforms relevant multi-shot generation approaches across all eight quantitative metrics and four LLM-based evaluations. This demonstrates its ability to create multi-shot cinematic videos with high fidelity to user instructions (OC, VTC), strong consistency across intra-shot and inter-shot (SC, BC, SC-E, BC-E, ISC), seamless cinematic transitions (TVS, STS), and superior overall quality (AQ, IQ, OVQ).

\begin{table}[t]
\caption{Percentage (\%) of user ratings in the four experiments of human evaluation for the results. }
\vspace{-4mm}
\begin{center}
{
    \setlength\tabcolsep{6pt}
    \centering
    \begin{adjustbox}{width={0.48\textwidth},totalheight={\textheight},keepaspectratio}
    \begin{tabular}{l | c c c c} \toprule
    % \multirow{1}{*}{Rating}
    Method  & VQE & TAE & SCE & ITE  \\ \midrule
    CineTrans~\cite{cinetrans} & $13.2$ & $18.4$ & $6.4$ & $16.4$   \\ 
    IC-LoRA~\cite{iclora} + WanX~\cite{wanx} & $24.4$ & $12.8$ & $13.2$ & $7.2$   \\ 
    StoryDiffusion~\cite{storydiffusion} + WanX~\cite{wanx} & $1.2$ & $8.0$ & $5.6$ & $3.6$   \\ 
    VideoGen-of-Thought~\cite{videogenofthought} & $2.8$ & $4.4$ & $0.4$ & $2.0$   \\ 
    MovieAgent~\cite{movieagent} & $0.8$ & $3.2$ & $1.6$ & $1.2$   \\ 
    Ours (STAGE) & $\mathbf{57.6}$ & $\mathbf{53.2}$ & $\mathbf{72.8}$ & $\mathbf{69.6}$   \\ 
        \bottomrule
\end{tabular}\label{tab:userstudy}
    \end{adjustbox}
    \vspace{-2mm}
}
\end{center}
\vspace{-4mm}
\end{table}

\begin{figure*}[t]
  \centering
  \includegraphics[width=\linewidth]{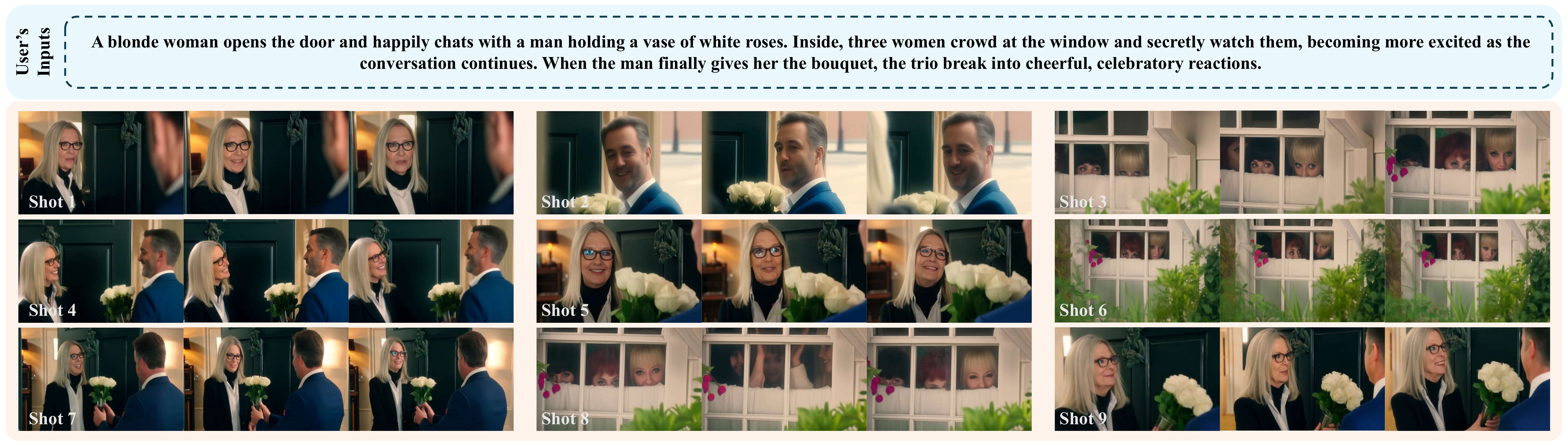}
  \caption{A long-range multi-shot video generation result of our STAGE framework, demonstrating high cross-shot consistency.}
  \label{fig:long_video}
  \vspace{-2mm}
  % \vspace{1mm}
\end{figure*}

\noindent \textbf{User study.}
In addition to qualitative and quantitative comparisons, we conduct four user studies to evaluate human preference:
\textit{(i)} \textbf{Visual Quality Evaluation (VQE):}
Participants are shown the generated results produced by STAGE and relevant multi-shot video generating methods. They are asked to select the most visually pleasing video.
\textit{(ii)} \textbf{Text Alignment Evaluation (TAE):} 
Given a corresponding text description, participants are instructed to select the video from the same set of generating results that best matches the description.
\textit{(iii)} \textbf{Shot Consistency Evaluation (SCE):} 
This study measures intra- and inter-shot visual consistency in multi-shot videos. Participants evaluate which sequence maintains stable appearance, motion, and spatial coherence across frames and shots.
\textit{(iv)} \textbf{Inter-shot Transition Evaluation (ITE):} 
This evaluation measures perceptual continuity and narrative logic between shots, where participants select the video with smoother visual transitions and more natural scene progression.
For each experiment, we randomly select 20 samples from the dataset, and recruit 25 volunteers from Amazon Mechanical Turk (AMT) to provide independent evaluations. As shown in \cref{tab:userstudy}, our model achieves the highest preference scores in all experiments.

\subsection{Ablation study}
We discard several modules and establish three baselines to study the impact of the corresponding modules. The evaluation scores and visual results of the ablation study are presented in \cref{tab:comparison} and \cref{fig:ablation}, respectively.

\noindent \textbf{W/o Multi-shot Memory Pack (MMP).} 
We discard the multi-shot memory pack, which causes the model to be unable to obtain context from preceding shots. Therefore, the model struggles to maintain visual consistency across the generated sequence (lower SC-E and BC-E scores. As shown in the first row of \cref{fig:ablation}, the character's appearance changes, and the scene illogically shifts from night to day across the first shot and the third shot).

\noindent \textbf{W/o Dual-Encoding Strategy (DES).} 
We remove the dual-encoding strategy and generate the first and last frames independently. Lacking mutual reference, the start and end frames fail to share context, which leads to degraded intra-shot consistency (lower SC and BC scores. As shown in the second row of \cref{fig:ablation}, the castles generated in the start and end frames are inconsistent).

\noindent \textbf{W/o Two-stage Training Scheme (TTS).}  
We discard the two-stage training scheme, training the model solely through supervised fine-tuning. Lacking exposure to negative examples, the model struggles to capture valid inter-shot transitions, leading to motion discontinuities (lower TVS score. As shown in the third row of \cref{fig:ablation}, an abrupt cut where the man is looking down in the first shot, but the camera is already shooting the sky in the second shot).

\subsection{Long multi-shot video generation}
As shown in \cref{fig:long_video}, we demonstrate STAGE's capability to generate long multi-shot videos by composing individual shots into a coherent and extended narrative.

\section{Conclusion}
In this paper, we introduced STAGE, a novel storyboard-anchored workflow for generating coherent multi-shot videos. Our key insight is to reformulate the task as a start-end frame pair prediction problem. This allows our proposed model (STEP$^2$) to effectively ensure cross-shot consistency and narrative structure using a multi-shot memory pack and a dual-encoding strategy. To capture nuanced cinematic language, we employ a two-stage training scheme with preference alignment, supported by our newly constructed ConStoryBoard dataset. Extensive experiments show that STAGE significantly outperforms state-of-the-art methods in narrative controllability, consistency, and cinematic quality. We believe STAGE is a significant step towards practical and user-directable video generation for long-form storytelling and AI-assisted filmmaking.

\noindent \textbf{Limitations.}
While multi-shot memory pack ensures long-range shot consistency with start-end frame pairs, the reliance on an off-the-shelf video generator for infilling can introduce temporal inconsistencies within the generated clips.
In future work, we will address this limitation by incorporating our multi-shot memory pack and structural storyboard into an end-to-end video generation backbone, enforcing dense consistency across all generated frames.

{
    \small
    \bibliographystyle{ieeenat_fullname}
    \bibliography{main}
}

\end{document}